\pdfoutput=1

\documentclass[11pt, dvipsnames]{article}

\renewcommand{\footnotesize}{\small}

\newcommand{\lspecialcell}[2][l]{%
  \begin{tabular}[#1]{@{}l@{}}#2\end{tabular}}

\usepackage{acl}

\usepackage{times}
\usepackage{latexsym}

\usepackage[T1]{fontenc}

\usepackage[utf8]{inputenc}

\usepackage{microtype}
\usepackage{xcolor}
\usepackage{amssymb}
\usepackage{tipa}
\usepackage{textcomp}
\usepackage{hyperref}
\usepackage{graphicx}
\usepackage{amsmath}
\usepackage{multirow}
\usepackage{cellspace}
\title{Mitigating Gender Bias in Machine Translation through\\ Adversarial Learning}

\author{Eve Fleisig \\
  UC Berkeley \\
  \texttt{efleisig@berkeley.edu} \And Christiane Fellbaum \\
  Princeton University \\
  \texttt{fellbaum@princeton.edu}}

\begin{document}
\maketitle
\begin{abstract}
Machine translation and other NLP systems often contain significant biases regarding sensitive attributes, such as gender or race, that worsen system performance and perpetuate harmful stereotypes. Recent preliminary research suggests that adversarial learning can be used as part of a model-agnostic bias mitigation method that requires no data modifications. However, adapting this strategy for machine translation and other modern NLP domains requires (1) restructuring training objectives in the context of fine-tuning pretrained large language models and (2) developing measures for gender or other protected variables for tasks in which these attributes must be deduced from the data itself.

We present an adversarial learning framework that addresses these challenges to mitigate gender bias in seq2seq machine translation.
Our framework improves the disparity in translation quality for sentences with male vs. female entities by 86\% for English-German translation and 91\% for English-French translation, with minimal effect on translation quality. The results suggest that adversarial learning is a promising technique for mitigating gender bias in machine translation.

\end{abstract}

\section{Introduction}
To avoid perpetuating harm, recent research has begun to examine how biases in NLP systems could be measured and reduced. Efforts to mitigate biases that rely on extensive dataset curation may be infeasible in some applications, such as translation of low-resource or morphologically complex languages. However, recent work suggests that adversarial learning can help to mitigate biases during training without the need to provide additional unbiased data or restructure the original model \cite{zhang2018mitigating}.\footnote{This approach is sometimes referred to as "adversarial debiasing," but following the authors themselves, we use "adversarial bias mitigation" to avoid the implication that all forms of bias are completely removed.} The method has shown promise in simple proof-of-concept applications, such as mitigating bias in word embeddings for use in analogies.

\begin{figure}
    \centering
    \includegraphics[width=0.34\textwidth]{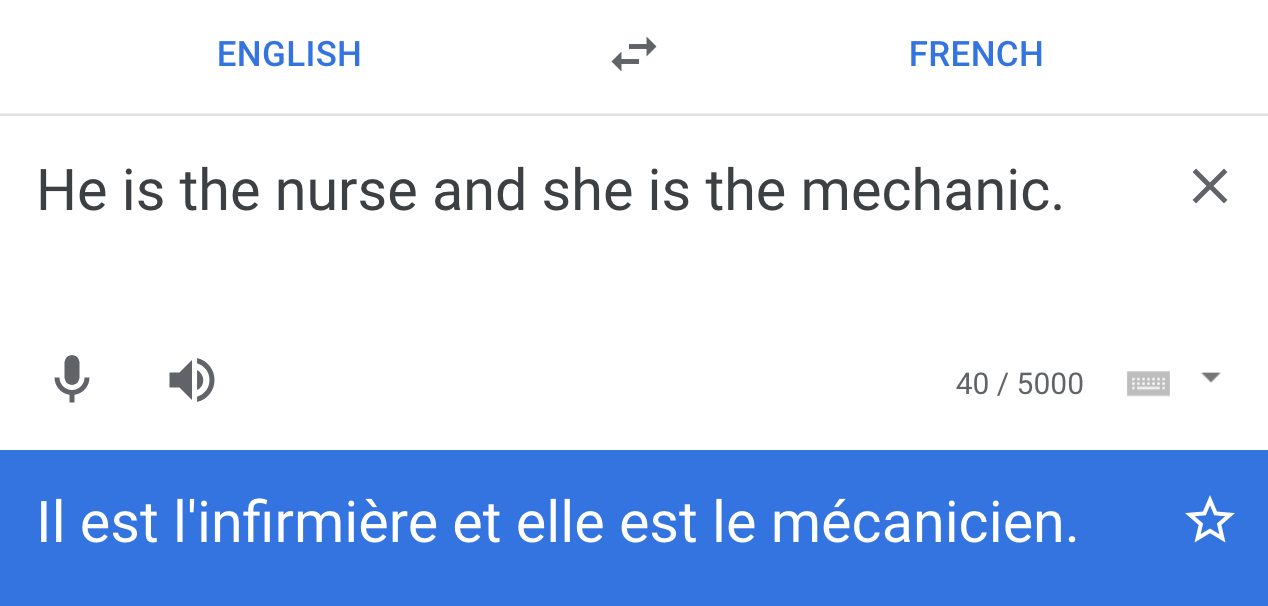}
    \caption{Example of gender bias in English-French translation using Google Translate. The system translates "nurse" to "l'infirmière," a female nurse, and "mechanic" to "le mécanicien," a male mechanic.}
    \label{fig:enfr-bias}
\end{figure}

Large language models, pre-trained without supervision and then fine-tuned for specific applications, have become a dominant paradigm in NLP. However, using adversarial learning for bias mitigation in such frameworks raises several questions.  How can we define a protected variable in the context of these models? How can we apply an adversarial framework for bias mitigation to a pre-training/fine-tuning setup? Finally, how can we quantitatively evaluate the extent to which this method mitigates gender bias?

We present a framework for leveraging adversarial learning to mitigate gender bias in machine translation that advances solutions to several issues faced when using this framework in modern NLP domains: we propose two measures for gender as a protected variable in the context of large language models, discuss how an adversarial framework can be applied during model fine-tuning, and present quantitative results on the effectiveness of this method at mitigating gender bias in machine translation. Our model reduces translation gender bias in the model T5 with little to no adverse effect on translation quality.

\section{Background and Related Work}

Recent work in the NLP community has stressed the need for studies of bias in NLP systems to discuss the normative reasoning behind why, how, and to whom an NLP system is harmful and ground this research in the literature outside NLP that examines how social processes lead to inequity \cite{blodgett-etal-2020-language}. \citet{beukeboom2019stereotypes} define linguistic bias as a "systematic asymmetry in language choice" that reflects stereotypical beliefs about social categories, as applied to either the category as a whole or its members. Under their Social Categories and Stereotypes Communication (SCSC) framework, these stereotypes skew perception of others by (1) preventing members of a social category from being viewed as distinct individuals (perceived entitativity), (2) reinforcing expectations about the social category, and (3) implying that characteristics are immutable and inherent to the group (perceived essentialism). Overtly or implicitly, stereotypes threaten or demean their targets. As a result, cognitive biases harm stereotyped individuals by causing people to fulfill stereotypical expectations, lowering their self-esteem, barring access to opportunities, and harming their mental and physical health \cite{beukeboom2019stereotypes}.

One way in which language encodes gender stereotypes is through the use of gendered terms. For example, studies examining job advertisements for male-dominated occupations found that female applicants were not only judged a poorer fit, but were also less likely to apply when a position was advertised in a masculine form (e.g., "chairman") versus a gender-neutral form (e.g., "chairperson") \cite{menegatti2017gender}. These effects could be particularly strong in languages with gender inflection, where most terms for professions have different forms depending on the person's gender (e.g., \textit{infirmier/infirmière} for ``nurse''). Thus, biases in NLP systems are destructive because they reproduce and reinforce pernicious societal power structures. Interventions in NLP that combat these biases present an opportunity to create more ethical and equitable systems that benefit all members of society.

\subsection{What Constitutes a Biased Translation?}

The harms of gender-stereotypical translations take the form of representation bias (misrepresenting a social category) and allocation bias (decreased performance for that social category) \cite{crawford2017}. Allocation bias with respect to gender in machine translation can occur when the accuracy of translation decreases according to a linguistic bias. This includes: (1) mistranslating sentences when they contain a female entity, but not when they contain a male  entity and (2) mistranslating sentences when they contain a counter-stereotypical association (such as a female doctor or male nurse, as in Figure \ref{fig:enfr-bias}), but not when they contain a stereotypical association. Mistranslations of sentences that contain a counter-stereotypical association (e.g., a female mechanic) simultaneously display allocation bias, because they fail to provide equal performance to different genders, and representational bias, because they reinforce gendered stereotypes. 

This research aims to minimize allocative and representational bias perpetuated \textit{within} a machine translation system as measured by the failure to meet a statistical fairness criterion.  Statistical fairness criteria that have been proposed include \textit{demographic parity}, \textit{equality of odds}, and \textit{equality of opportunity}  \cite{hardt2016equality, beutel2017data}; we use demographic parity, which defines a fair classifier as one in which predictions $\hat{Y}$ and the protected variable $Z$ are independent. That is,
$$P(\hat{Y} = \hat{y}) = P(\hat{Y} = \hat{y} | Z=z)$$
The adversarial method for bias mitigation used in this paper can be quickly extended to work with equality of odds and equality of opportunity (see Section \ref{section:approach}).

\subsection{Documenting Bias}

\citet{caliskan2017semantics} found that word embeddings exhibited gender and racial bias similar to those exhibited by humans and that machine translation systems exhibited gender bias in its translation of pronouns; subsequent studies found similar biases across other NLP tasks \cite{may2019measuring,zhao2017men,rudinger2018gender}. The translation biases found by \citet{caliskan2017semantics} raised awareness of bias in machine translation, leading some translation systems to introduce limited gender-specific translations as recently as 2020. However, more recent studies by \citet{kocmi-etal-2020-gender} and \citet{stanovsky2019evaluating} found that evidence of gender bias persisted across 10 languages over a total of 23 translation systems, including Google Translate, Microsoft Translator, Amazon Translate, and Systran.

Biases can be incorporated into machine learning systems during different stages of model development. Stereotyped associations and unbalanced representation of different demographics in training corpora (``dataset bias''), along with bias amplification effects during model training, result in models that exhibit biases far beyond real-world disparities \cite{rudinger2018gender, lu2018gender}.

\subsection{Mitigating Bias}
\citet{font-costa-jussa-2019-equalizing} propose reducing gender bias in English-Spanish machine translation by adjusting word embeddings and suggest that this method improves translation gender bias on some examples. Meanwhile, several efforts at mitigating translation bias have intervened through dataset curation, either by refining or annotating existing training sets or creating new datasets for fine-tuning. \citet{vanmassenhove2019getting} tagged sentences with information on the speaker's gender, which affects the grammatical gender of words in some languages and may inform word choice more generally, to improve the translation quality of sentences spoken by women. \citet{saunders-etal-2020-neural} and \citet{stafanovics-etal-2020-mitigating} similarly use training data annotated with gender tags. \citet{saunders2020reducing} addressed gender bias using corrective fine-tuning with a smaller, handcrafted dataset of gender-balanced sentences and suggested methods for swapping the genders of entities in languages with gender inflection. The authors note that there is usually a tradeoff between bias mitigation on the WinoMT dataset (see Section \ref{section:approach}) and translation quality.

However, data-centric approaches to bias mitigation suffer from the difficulty of collecting data for specific applications, particularly for low-resource and morphologically complex languages. In addition, approaches that intervene before training may address dataset bias but not bias amplification. In the case of gender bias, data-centric approaches may also be more difficult to apply to languages with grammatical gender, in which gender may be represented not only in pronouns or occupations, but also through the inflection of nouns, adjectives, and other parts of speech \cite{zmigrod-etal-2019-counterfactual}.

\subsection{Adversarial Learning Approaches} \label{subsection:adversary-papers}
\citet{zhang2018mitigating} proposed an adversarial technique for general bias mitigation during training. An adversary is trained to predict a \textit{protected variable} (i.e., gender), while the model learns to prevent the adversary from predicting the protected variable (see Section \ref{section:approach}). They applied adversarial bias mitigation to two proof-of-concept tasks: income prediction on the UCI Adult dataset and analogy completion (e.g., ``man : woman :: king : \_\_''). For the analogy completion task, the study defined gender bias according to the notion of a ``gender direction'' proposed by
\citet{bolukbasi2016man}. This method measures gender bias as the magnitude of the projection $proj_g y$ of a sentence $y$ onto the ``gender direction'' $g$ of a word embedding space.

Zhang et al. found that the method substantially reduced bias in the income prediction task. They also gave examples of bias reduction in the analogy completion task, such as a decreased likelihood of choosing ``nurse'' as the female equivalent of ``doctor''; however, they do not provide  evidence of systematic bias mitigation in the model overall, possibly due to the scarcity of datasets for testing gender bias at the time of the study's publication. Thus, the exact degree to which this method can mitigate bias remained an open question.

\citet{kumar2019} independently introduced an adversarial framework for text classification to prevent confounding variables, such as the mention of a particular country, from overly affecting classification, such as language identification. \citet{xia2020demoting} drew on this vein of research to mitigate racial bias in a small LSTM-based hate speech detection model, using tweets that were pre-annotated with the probable race of the author. However, previous work on adversarial bias mitigation has yet to examine issues pertaining to measuring gender as a protected variable when not prelabeled or apply the adversarial technique to large language models.

\section{Approach}
\label{section:approach}
The adversarial framework for bias mitigation has several advantages that make it suitable for machine translation and LLM-based tasks more broadly. Adversarial bias mitigation is a model-agnostic strategy: so long as the model trains using gradient descent, the complexity of the model being trained does not affect the overall framework. This advantage makes it suitable for mitigating biases under the common framework of pre-training a large language model on a general language understanding task, then fine-tuning on machine translation or other specific applications. By modifying the training process itself, it also works to mitigate the effects of bias amplification as well as dataset bias. In addition, unlike bias mitigation techniques that intervene before or after the training process, adversarial bias mitigation does not require extensive modifications to the training data or additional data collection, which makes it easier to extend to new tasks or  low-resource domains.

\subsection{Framework}
We use an adversarial approach that mitigates gender bias by defining a training objective that encourages a model to minimize the gendered information encoded in output sentence embeddings beyond what is strictly necessary to translate the sentence correctly. In this approach, a prediction model $M$ with weights $W$ learns to predict an output $Y$ from input $X$ while remaining neutral with respect to the protected variable $Z$. The adversary $A$ attempts to predict $Z$ from the model's output predictions $\hat{Y}$. Then, $W$ is updated according to:
$$\nabla_W L_P - \text{proj}_{\nabla_W L_A} \nabla_W L_P - \alpha \nabla_W L_A$$

\noindent where $\alpha$ is a tuneable hyperparameter. This training objective penalizes the prediction model for helping the adversary to determine the value of the protected variable \cite{zhang2018mitigating}.

In this work, we define fairness through demographic parity (see Section \ref{subsection:adversary-papers}). Replacing the objective of demographic parity used in this paper with either equality of odds (conditional independence between $\hat{Y}$ and $Z$, given $Y$) or equality of opportunity for a group $y$ (independence between $\hat{Y}$ and $Z$, conditioned on $Y=y$) requires minimal changes: for equality of odds, the adversary can be given access to the target translation $Y$ as well as the prediction $\hat{Y}$; for equality of opportunity on a class $y$, the adversary should only train on examples for which $Y=y$.

\subsection{Defining the Protected Variable Z}
\subsubsection{Method 1: Gender Direction from Sentence Encodings}

To define the protected variable $Z$, we extend the notion of a "gender direction" $g$ from \citet{bolukbasi2016man} and \citet {zhang2018mitigating}. Zhang et al. defined 10 male/female word pairs (e.g., he/she, him/her), and from these defined a \textit{bias subspace}, the space spanned by the top principal component of the differences. The unit vector $g$ representing the bias subspace thus approximates the "she-he direction" of the word embedding space. They then defined the protected variable for the task of completing analogies based on word embeddings as $proj_g y$, the projection along the gender direction of the word $y$ that completes the analogy.

Extending this formulation to sentence embeddings, we calculate the bias subspace from the top principal component of the model $M$'s sentence encodings for the 10 male/female word pairs to find $g$ for the sentence embedding space of the output of the model.\footnote{$g$ is reduced to 30,000 entries to prevent the principal component analysis from becoming prohibitively expensive.} We then define $Z$ as $proj_g y$, the projection of the sentence encoding along the gender direction. After masking all pronouns in the model’s predicted translation $\hat{y}$ of a sentence, the adversary attempts to predict $proj_g \hat{y}$, while the model is trained to avoid providing information that allows the adversary to do so.\footnote{First names were uncommon enough in the data that we found masking them was not needed for the method to work.}

\subsubsection{Method 2: Pronoun Usage Heuristic}
An open question is whether there are ways of defining the protected variable $Z$ that are more effective at mitigating bias or  otherwise useful for NLP tasks. Thus, we also implemented a \textit{pronoun usage heuristic} for defining the protected variable $Z$. Under this alternative metric, $Z$ is defined as 1 if a feminine pronoun occurred, -1 if a masculine pronoun occurred, and 0 if both occurred. This simpler metric has some advantage in terms of computational cost, since the principal component analysis and matrix multiplications needed to calculate the projection of each encoded sentence on $g$ require some expensive calculations before training. On the other hand, evidence that the gender direction metric is more effective than the pronoun usage metric would indicate that calculating $Z$ from how the model encodes the sentence provides additional information that is useful for mitigating bias in that model.

\begin{figure}
    \centering
    \includegraphics[width=0.31\textwidth]{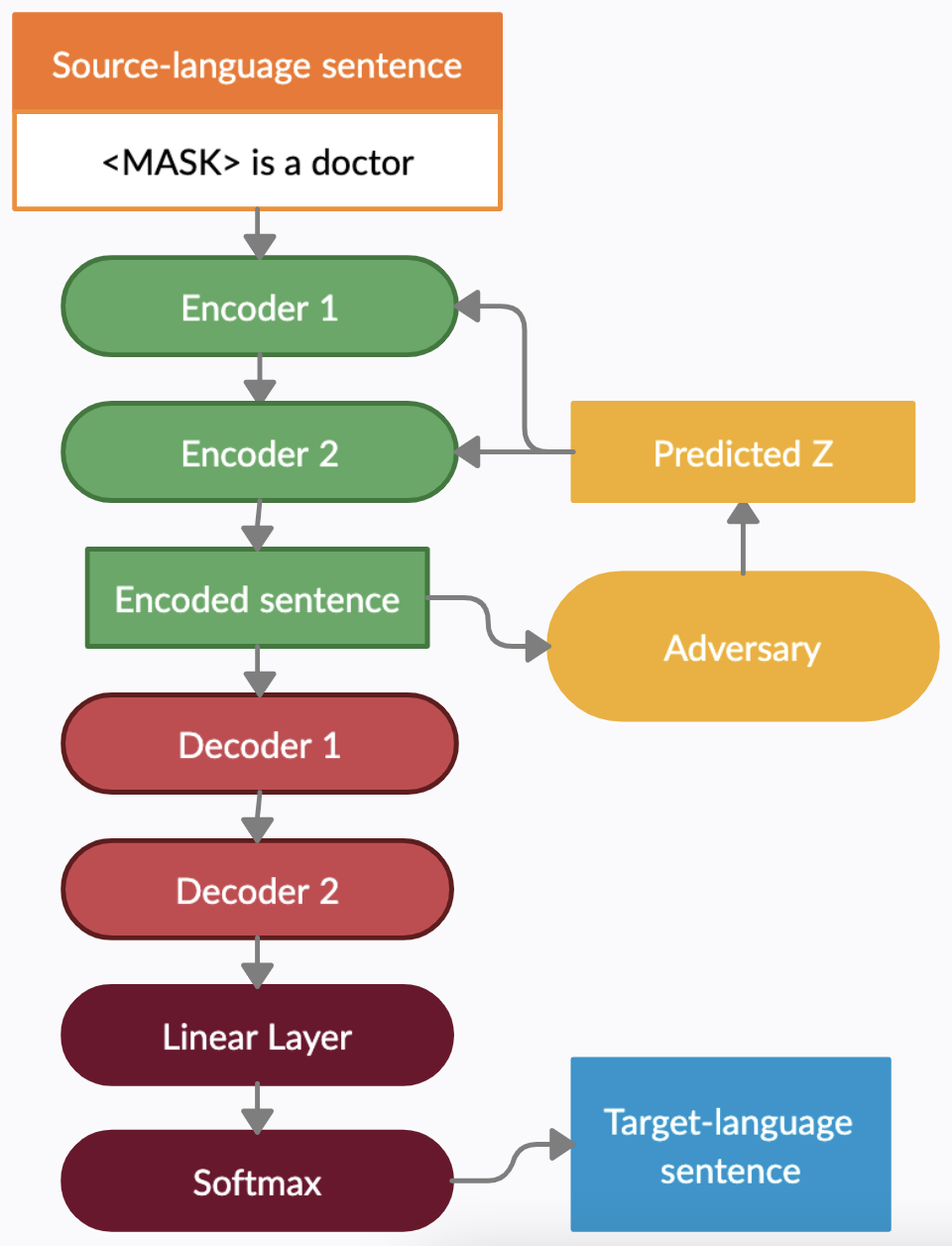}
    \caption{Our framework for adversarial bias mitigation in machine translation with T5.}
    \label{fig:flow}
\end{figure}

\section{Implementation} \label{section:implementation}
We fine-tuned the model T5 \cite{raffel2019exploring} on English-French and English-German translation under our adversarial bias mitigation paradigm.\footnote{In an effort to examine performance on low-resource languages, we also performed initial experiments on English-Czech translation with an order of magnitude smaller dataset; however, issues with translation quality suggest more extensive modifications are necessary to extend this technique to low-resource settings.} The adversarial intervention occurred during fine-tuning alone, without intervening during the pretraining stage. Since T5 is an encoder-decoder model, $\hat{Y}$ (the representation of the encoded sentence) is the output of the second encoder of T5. For the gender direction method, the protected variable $Z$ for a sentence $S$ was found through principal component analysis on $\hat{Y}$ on the pretrained model before fine-tuning. During training, $\hat{Y}$ is then sent to the adversary $A$, which attempts to predict $Z$ (Figure \ref{fig:flow}).

We used the WMT-2014 corpus \cite{bojar-etal-2014-findings} to train the model and evaluate for translation quality (see Section \ref{section:results}). For each translation pair, the model was fine-tuned on a subset of 100,000 examples that contained at least one gendered pronoun. This was done to ensure that the training set included a higher proportion of sentences with gendered entities, since the majority of sentences in the original dataset contained no gendered entities at all. We masked all source sentences' gendered pronouns in the training data.

The development and test sets each consisted of 50,000 random unseen sentence pairs from the corpus, including sentences without gendered pronouns, to ensure that the intervention during fine-tuning did not hinder the model's ability to translate in general. (See Appendix \ref{app:implementation} for hyperparameter details.)

\begin{figure*}[h]
    \centering
    \includegraphics[width=0.33\textwidth]{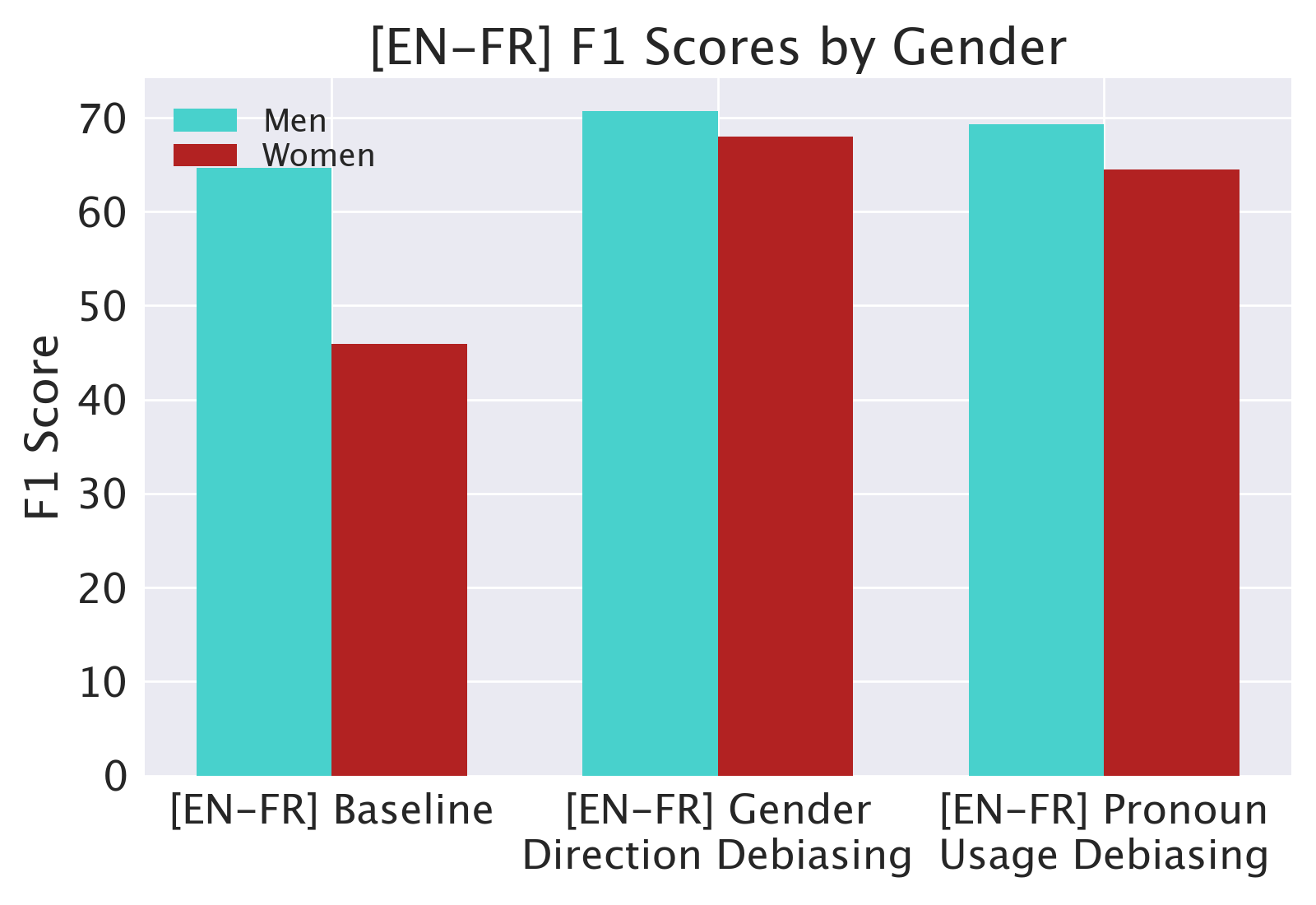} \includegraphics[width=0.325\textwidth]{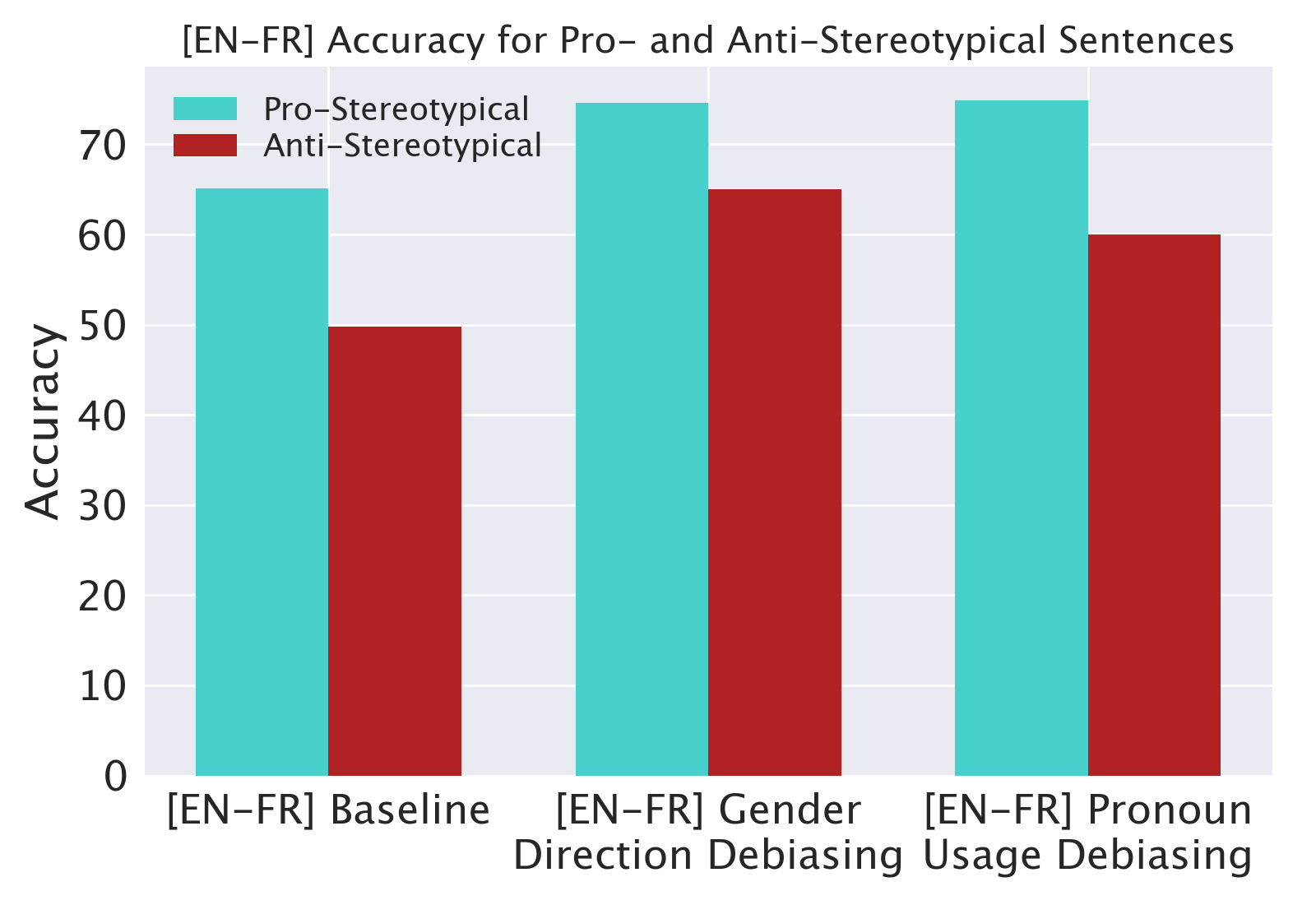} \includegraphics[width=0.325\textwidth]{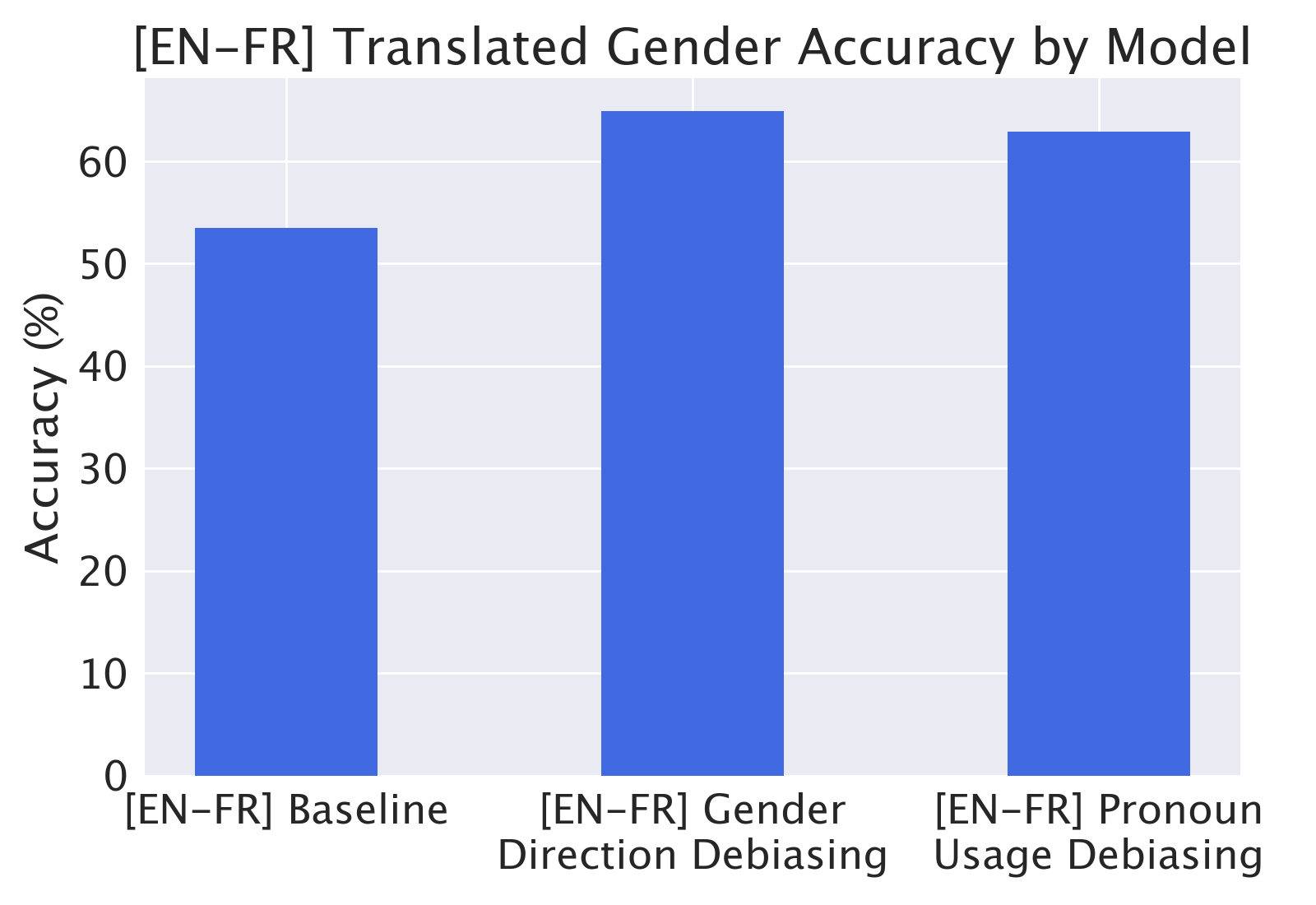}

    \includegraphics[width=0.33\textwidth]{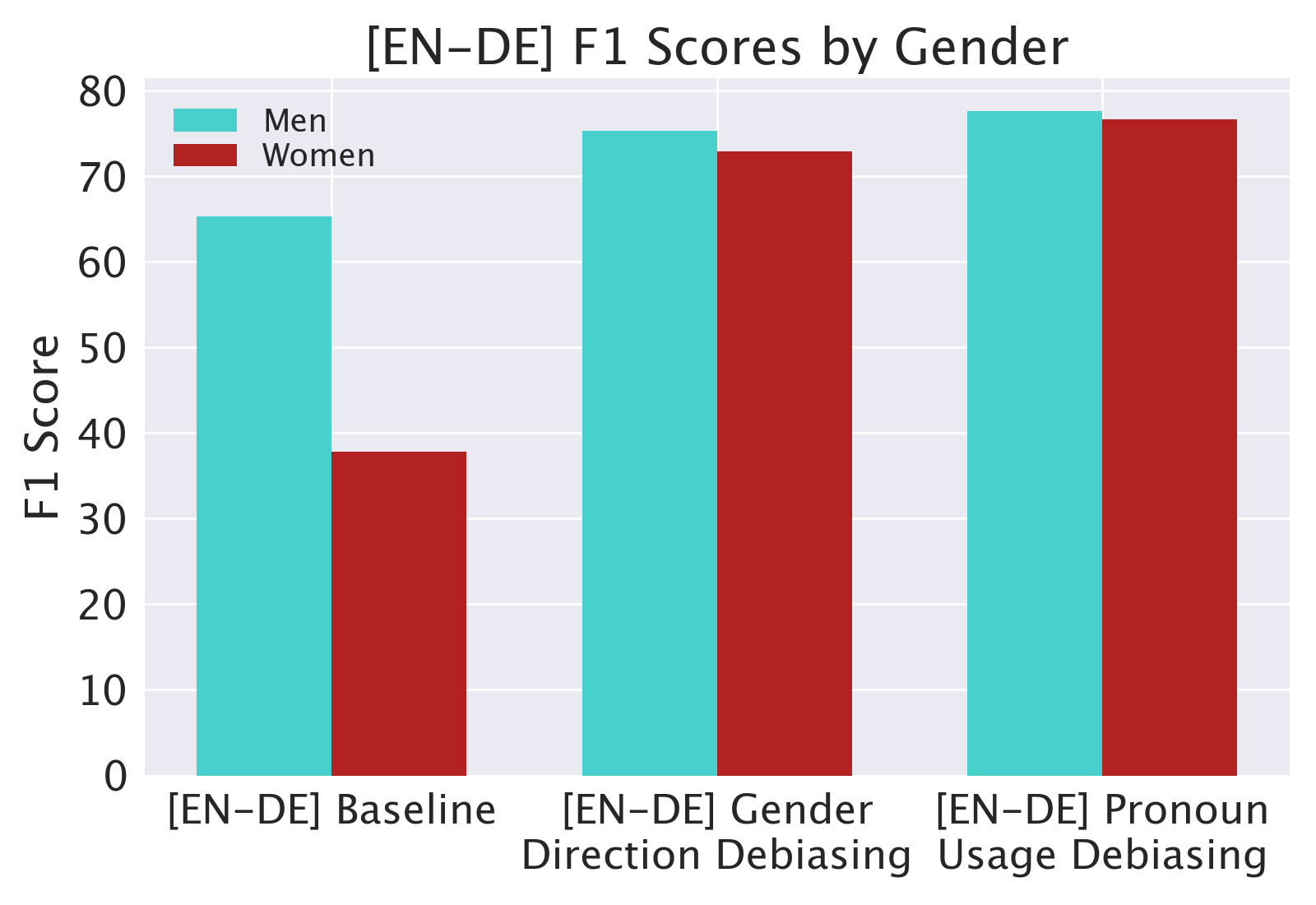} \includegraphics[width=0.325\textwidth]{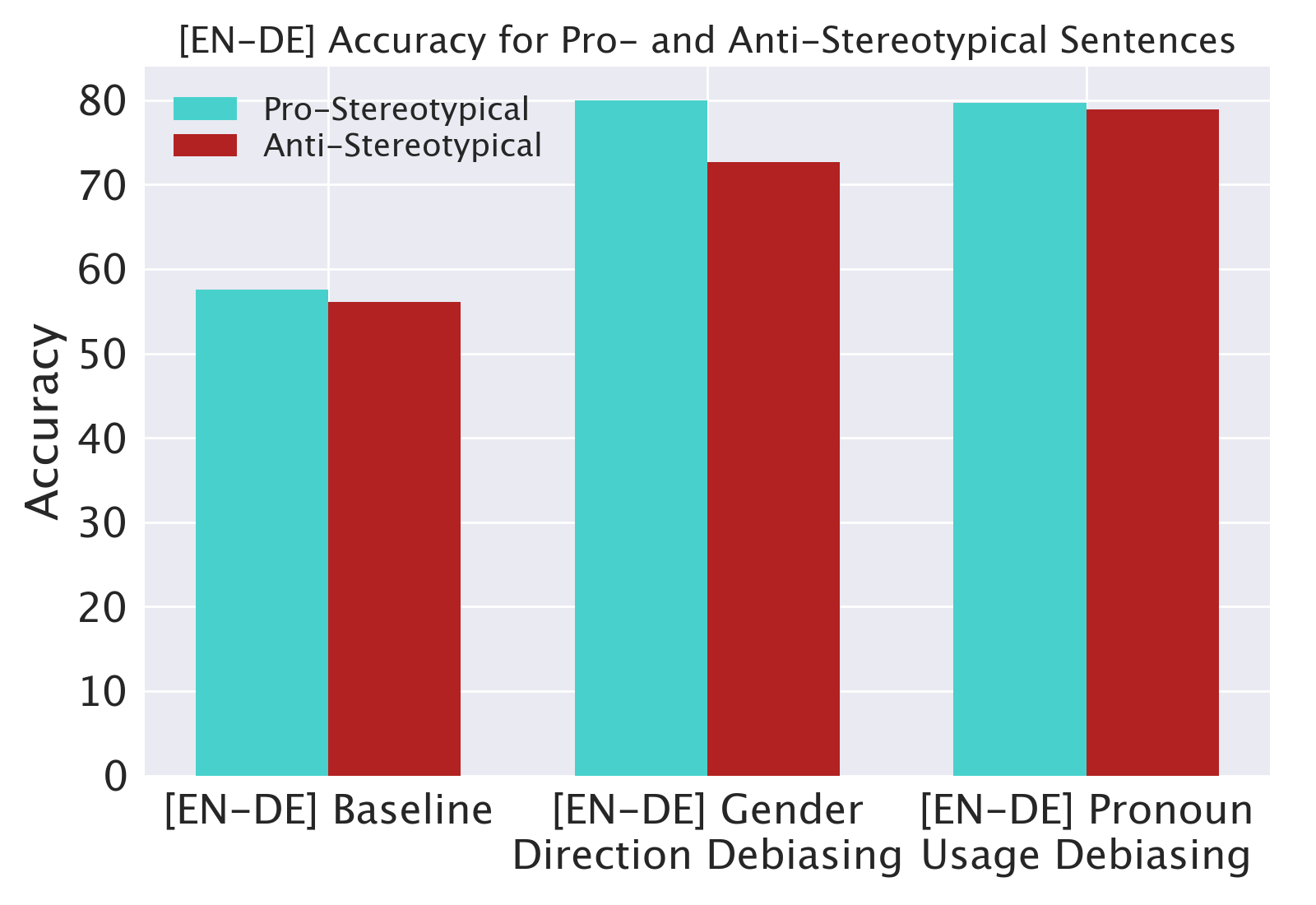} \includegraphics[width=0.325\textwidth]{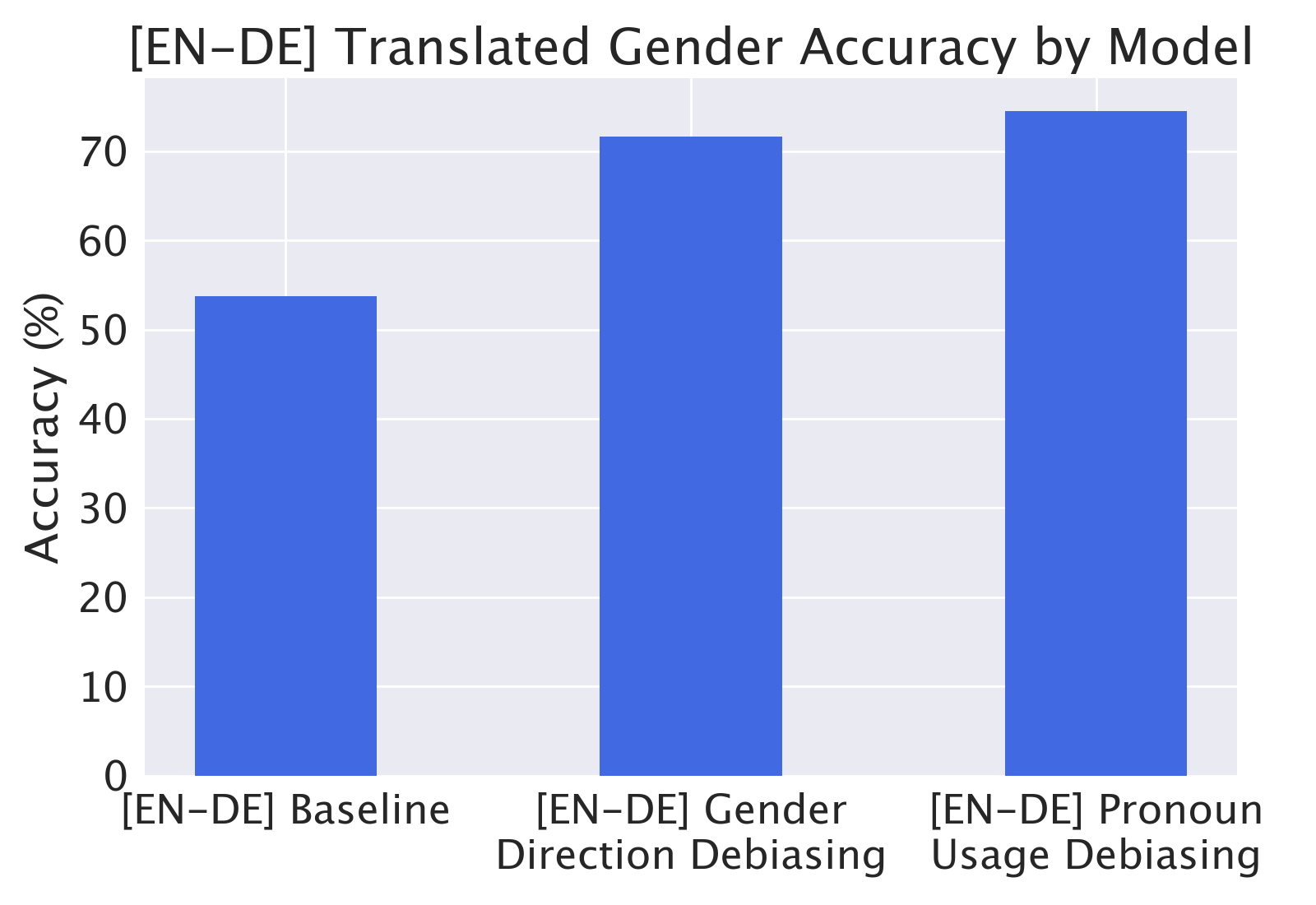}
    \caption{\textbf{Left}: Difference in F1 scores on  WinoMT for sentences involving male vs. female entities.\\ \textbf{Middle}: Accuracy scores on WinoMT dataset for sentences involving pro- vs. anti-stereotypical role assignments.\\ \textbf{Right}: Overall accuracy of preserving the gender of an entity upon translation in WinoMT.}
    \label{fig:acc}
\end{figure*}

\section{Evaluation Results}\label{section:results}
The model was tested on both translation accuracy (BLEU score on the WMT-2014 data) and minimization of bias. For bias mitigation, the model was tested on the WinoMT dataset \cite{stanovsky2019evaluating}, a challenge set for gender bias in machine translation consisting of sentences balanced between male and female genders and between male and female gender role assignments (e.g. male doctor/female doctor, female nurse/male nurse).

The primary metrics used to measure performance on the WinoMT dataset are  $\Delta G$, $\Delta S$, and the overall accuracy of preserving the gender of an entity upon translation ("acc."). $\Delta G$ denotes the difference in F1 scores between sentences involving masculine entities and sentences involving feminine entities. $\Delta S$ denotes the difference in accuracy on correctly translating the antecedent between sentences involving pro-stereotypical (e.g., male doctor/female nurse) and anti-stereotypical (e..g, female doctor/male nurse) role assignments.

\subsection{Results on Bias Mitigation}
Both methods of bias mitigation significantly reduced gender bias in the translated output, with minimal change in translation accuracy (Table \ref{table:summary}). The disparity in F1 scores between sentences involving male and female entities ($\Delta G$) decreased from 18.8 (EN-FR) and 27.5 (EN-DE) in the baseline to 4.8 and 10.0 under the pronoun usage heuristic and to 2.0 and 2.4 using the gender direction (86\% and 91\% relative improvement, respectively).

The accuracy of preserving the genders of entities in translation increased from 53.5\% and 53.7\% in the baseline to 62.9\% and 78.9\% under the pronoun usage heuristic and to 64.9\% and 72.7\% using the gender direction (Figure \ref{fig:acc}). The disparity in accurate translation of antecedents in sentences with stereotypical and reverse-stereotypical role assignments ($\Delta S$) decreased in most cases, from 15.3 and 1.5 in the baseline to 14.9 and 0.8 under the pronoun usage heuristic and to 9.0 and 7.3 using the gender direction. \footnote{The WinoMT dataset for gender bias evaluation is fairly new, which places a limitation on comparing the results on these evaluation metrics to previous studies of bias mitigation in machine translation. One study whose evaluation bears some similarity is by \citet{stafanovics-etal-2020-mitigating}, who added target gender annotations to WMT-2015 to mitigate bias in the Sockeye MT model. Their model's performance went from a baseline of $\Delta G = 29.8, \Delta S = 11.8$ (EN-FR) and $\Delta G = 10.2, \Delta S = 14.4$ (EN-DE) to $\Delta G = 1.6$ and $\Delta S = 10.1$ (EN-FR) and $\Delta G = -4.7, \Delta S = 1.7$ (EN-DE).}
(The greater effect on $\Delta G$ compared to the other metrics is consistent with other studies on the WinoMT dataset, in which $\Delta G$ usually displays the most visible change after bias mitigation \cite{stafanovics-etal-2020-mitigating, kocmi-etal-2020-gender}.)

\begin{table*}
\centering
\resizebox{0.65\textwidth}{!}{
\begin{tabular}{Sl Sl Sl Sl Sl Sl}
\hline
    & & \textbf{BLEU} &
    $\Delta\mathbf{G}$ &
    \textbf{Acc.} &
    $\Delta \mathbf{S}$\\
    \hline
\multirow{3}{*}{EN-FR} &\textbf{Baseline} & 30.7  & 18.8 & 53.5 & 15.3\\
&\textbf{Gender direction method} & 29.2       &  \textbf{2.7}   & \textbf{64.9} & \textbf{9.0} \\
&\textbf{Pronoun usage method}    & \textbf{33.0}       & 4.8 & 62.9 & 14.9\\
\hline
\multirow{3}{*}{EN-DE} &\textbf{Baseline} & 28.4  & 27.5 & 53.7 & 1.5\\
&\textbf{Gender direction method} & \textbf{31.5}       &  \textbf{2.4}   & 72.7 & 7.3 \\
&\textbf{Pronoun usage method}    & 29.9       & 10 & \textbf{78.9} & \textbf{0.8}\\
\hline
\end{tabular}}
\caption{Results for bias mitigation and translation quality on all three models. BLEU scores were evaluated on the WMT-2014 data. $\Delta G$ denotes the difference in F1 scores on the WinoMT dataset between sentences with masculine entities and sentences with feminine ones. $\Delta S$ denotes the difference in accuracy scores on the WinoMT dataset between sentences with pro-stereotypical and anti-stereotypical role assignments.}
\label{table:summary}
\end{table*}

\subsection{Results on Translation Quality}
The translation quality, as measured by BLEU score, displayed only small variations under both bias mitigation methods. In English-French translation, BLEU score decreased slightly from 30.7 to 29.2 when using the gender direction, and in fact increased from 30.7 to 33.0 when using the pronoun usage heuristic. In English-German translation, the BLEU score increased in both cases, from 28.4 in the baseline to 31.5 when using the gender direction method and 29.9 with pronoun usage method. Since a slight decrease in translation quality might be expected when training with more complex objectives for bias mitigation, the increase in translation quality under most circumstances might suggest that the adversarial bias mitigation framework provides some additional benefit in terms of avoiding overfitting: training the model to avoid gendered associations could permit it to generalize more easily to unseen data.

\begin{table*}[h!]
\centering
\resizebox{0.97\textwidth}{!}{%
\begin{tabular}{Sl Sl Sl }
\textbf{Source sentence}                               &  \textbf{Translation (baseline)}                   &  \textbf{Translation (after bias mitigation)}\\
\hline
\lspecialcell{{\color{BurntOrange}\textbf{The developer}} argued with\\ the designer because {\color{BrickRed}\textbf{she}} did \\not like the design.}         & \lspecialcell{{\color{Cerulean}\textbf{Le promoteur}} a eu des \\échanges avec l’auteur parce \\qu’{\color{Cerulean}\textbf{il}} n’a pas aimé le dessin. }                               & \lspecialcell{{\color{BrickRed}\textbf{La développeure}} a contesté la conception\\ parce qu’{\color{BrickRed}\textbf{elle}} n’aimait pas le design. \\\textit{(gender direction)}} \\
\hline
\lspecialcell{The physician hired {\color{BurntOrange}\textbf{the}} \\{\color{BurntOrange}\textbf{secretary}} because {\color{Cerulean}\textbf{he}} was \\highly recommended.}                & \lspecialcell{Le médecin embaucha {\color{BrickRed}\textbf{la}} {\color{BrickRed}\textbf{secrétaire}}\\parce qu'{\color{BrickRed}\textbf{elle}} était recommandée \\à la haute recommandation.}                 & \lspecialcell{Le médecin a embauché {\color{Cerulean}\textbf{le}} {\color{Cerulean}\textbf{secrétaire}} \\parce qu'{\color{Cerulean}\textbf{il}} était hautement recommandé.\\ \textit{(pronoun usage)}} \\
\hline
\end{tabular}}
\caption{Sample translations by the baseline model and by the models trained with each of the bias mitigation methods. Colors indicate entities' grammatical genders: {\color{Cerulean}\textbf{blue}}  for male entities and pronouns, {\color{BrickRed}\textbf{red}} for female ones, and {\color{BurntOrange}\textbf{orange}} for neutral ones.}
\end{table*} \label{table:examples}

The gender direction method provided the best results in terms of bias mitigation on English-French translation; in English-German translation, the pronoun usage method was more suited to removing stereotypical associations, whereas the gender direction method was more suited to improving performance on sentences with female entities. Both methods performed comparably in terms of translation accuracy, though the pronoun usage method provided a significantly greater boost in English-French translation. Linguistic differences between French and German, such as the existence of a neuter gender in German, could account for these differences; future work could examine more languages across of a variety of language families to further understand this behavior. In addition, these results suggest that when choosing methods for bias mitigation, there may be a tradeoff, albeit small, between the best possible fairness and the best possible accuracy. For example, in English-French translation, the pronoun usage method provides the best improvement in overall translation quality, while the gender direction method is best at improving gender bias across all metrics.

The particular goals of the application for which a translation model is deployed could determine which method of measuring gender bias is best for that context: the gender direction method provides more dramatic improvements in some cases, whereas the pronoun usage method provides more consistent bias mitigation across all metrics in both languages. That said, both methods of measuring gender bias in conjunction with adversarial learning resulted in significant decreases in gender biases across nearly all metrics.

\subsection{Examples of Bias Mitigation}

Table \ref{table:examples} gives examples of translations with and without different bias mitigation techniques on the WinoMT dataset. Sentences in the WinoMT dataset are designed such that the model must use context to determine which entity in a sentence (e.g., "the developer" or "the designer" in Table 1) corresponds to the gendered pronoun, since resolving the coreference to either entity would be syntactically correct. 

Without using bias mitigation, the baseline model translates both gendered entities and pronouns in stereotypical ways: a female developer becomes \textit{le promoteur}, the male developer, and a male secretary becomes \textit{la secrétaire}, the female secretary. The gendered pronouns associated with these entities are also translated to the stereotypical gender for those professions. Both methods of bias mitigation, however, translate both the gendered entities and the gendered pronouns that correspond to them correctly in these instances. The female developer becomes \textit{la développeure} and the correct pronoun \textit{elle} is used for her; the male secretary becomes \textit{le secrétaire}, using the correct pronoun \textit{il}. These differences illustrate how both the gender direction and pronoun usage methods can successfully mitigate bias through adversarial learning. 

\section{Conclusion}\label{section:conclusion}
Linguistic biases serve as a vehicle for harmful stereotypes that demean individuals and communities, harm their mental and physical health, and worsen life outcomes \cite{beukeboom2019stereotypes}. Recent studies indicate that NLP systems perpetuate these biases, reproducing stereotypes in their output and disproportionately producing demeaning or outright incorrect output for groups that face societal discrimination. For example, translation systems often translate pronouns or gender inflections incorrectly when they correspond to counter-stereotypical professions.

Adversarial bias mitigation intervenes during training by introducing an adversary that attempts to predict a protected variable from the output of the model. It is a model-agnostic strategy, permitting complex models to be substituted into the framework without changing the overall setup, which is convenient for pre-training/fine-tuning setups. It requires no modifications to the training data or additional data collection, permitting it to be extended to new tasks or low-resource domains more easily. By intervening during training itself, this approach also aims to mitigate both dataset bias and bias amplification.

In this work, we addressed  several open questions raised by previous research into the adversarial approach to bias mitigation: (1) how to define gender as the protected variable in more complex applications, particularly in language tasks where such information is not prelabeled; (2) how to apply the adversarial framework to a pre-training/fine-tuning setup with large language models, as has become the norm; and (3) whether the adversarial framework is indeed effective as measured by quantitative evaluations on realistic tasks.

We presented an adversarial framework for mitigating gender bias in machine translation. Our approach proposes two measures for gender as a protected variable in the context of large language models, the \textit{gender direction} method and \textit{pronoun usage} method. We then applied the adversarial framework to English-French and English-German machine translation. For both the gender direction and pronoun usage methods, the difference in F1 scores between sentences in the WinoMT dataset involving male and female entities decreased, and for the pronoun usage method, the difference in accuracy between pro- and anti-stereotypical sentences also decreased. In addition, the accuracy of preserving the gender of an entity upon translation increased and the accuracy of translating pro- and anti-stereotypical sentences increased for both methods. Furthermore, mitigating gender bias did not come at the expense of translation quality. In fact, translation accuracy slightly increased in most cases, suggesting that the method might provide some additional ability to generalize to new examples. 

The gender direction method was significantly more successful at mitigating bias in some cases, whereas the pronoun usage method  provided more consistent but usually less thorough bias mitigation. Nonetheless, both methods were effective at mitigating gender bias in machine translation. The results suggest that the adversarial framework is a promising technique for mitigating biases in common and complex NLP tasks.

\subsection{Future Work}
A broader avenue of research concerns extending the adversarial framework to other NLP tasks and to protected variables such as race or religion, for which indicators of the protected variable may be more difficult to measure. Possible extensions to this work within machine translation could examine the efficacy of this method on different translation pairs, especially between more dissimilar languages. Translation from languages with more complex systems of gender inflection might require more complex strategies for defining the protected variable tailored to their syntactic and morphological features. Another key direction is to account for nonbinary or transgender users and others who face unique forms of gender discrimination (e.g., misgendering) or for whom gender bias mitigation based on binary notions of gender would result in oversimplified interventions \cite{cao2019toward}.

Complementary research can involve integrating stakeholders in the development of NLP systems, such as by surveying users of various genders or collaborating with language reclamation activists. Allowing users to contest or modify the decisions made by a system, such as by allowing users to correct biased translations or choose between multiple translations, could also improve trust in the fairness of a translation system \cite{vaccaro2019contestability}. Combining multiple types of interventions can allow the NLP community to address these issues, which require both technical and ethical insights.

\bibliography{anthology,custom}

\begin{thebibliography}{27}
\expandafter\ifx\csname natexlab\endcsname\relax\def\natexlab#1{#1}\fi

\bibitem[{Beukeboom and Burgers(2019)}]{beukeboom2019stereotypes}
Camiel~J Beukeboom and Christian Burgers. 2019.
\newblock \href {10.12840/issn.2255-4165.017} {How stereotypes are shared
  through language: a review and introduction of the social categories and
  stereotypes communication ({SCSC}) framework}.
\newblock \emph{Review of Communication Research}, 7:1--37.

\bibitem[{Beutel et~al.(2017)Beutel, Chen, Zhao, and Chi}]{beutel2017data}
Alex Beutel, Jilin Chen, Zhe Zhao, and Ed~H. Chi. 2017.
\newblock \href {http://arxiv.org/abs/1707.00075} {Data decisions and
  theoretical implications when adversarially learning fair representations}.

\bibitem[{Blodgett et~al.(2020)Blodgett, Barocas, Daum{\'e}~III, and
  Wallach}]{blodgett-etal-2020-language}
Su~Lin Blodgett, Solon Barocas, Hal Daum{\'e}~III, and Hanna Wallach. 2020.
\newblock \href {https://doi.org/10.18653/v1/2020.acl-main.485} {Language
  (technology) is power: A critical survey of {``}bias{''} in {NLP}}.
\newblock In \emph{Proceedings of the 58th Annual Meeting of the Association
  for Computational Linguistics}, pages 5454--5476, Online. Association for
  Computational Linguistics.

\bibitem[{Bojar et~al.(2014)Bojar, Buck, Federmann, Haddow, Koehn, Leveling,
  Monz, Pecina, Post, Saint-Amand, Soricut, Specia, and
  Tamchyna}]{bojar-etal-2014-findings}
Ond{\v{r}}ej Bojar, Christian Buck, Christian Federmann, Barry Haddow, Philipp
  Koehn, Johannes Leveling, Christof Monz, Pavel Pecina, Matt Post, Herve
  Saint-Amand, Radu Soricut, Lucia Specia, and Ale{\v{s}} Tamchyna. 2014.
\newblock \href {https://doi.org/10.3115/v1/W14-3302} {Findings of the 2014
  workshop on statistical machine translation}.
\newblock In \emph{Proceedings of the Ninth Workshop on Statistical Machine
  Translation}, pages 12--58, Baltimore, Maryland, USA. Association for
  Computational Linguistics.

\bibitem[{Bolukbasi et~al.(2016)Bolukbasi, Chang, Zou, Saligrama, and
  Kalai}]{bolukbasi2016man}
Tolga Bolukbasi, Kai-Wei Chang, James Zou, Venkatesh Saligrama, and Adam Kalai.
  2016.
\newblock \href {https://dl.acm.org/doi/10.5555/3157382.3157584} {Man is to
  computer programmer as woman is to homemaker? debiasing word embeddings}.
\newblock In \emph{Proceedings of the 30th International Conference on Neural
  Information Processing Systems}, NIPS'16, page 4356–4364, Red Hook, NY,
  USA. Curran Associates Inc.

\bibitem[{Caliskan et~al.(2017)Caliskan, Bryson, and
  Narayanan}]{caliskan2017semantics}
Aylin Caliskan, Joanna~J. Bryson, and Arvind Narayanan. 2017.
\newblock \href {https://doi.org/10.1126/science.aal4230} {Semantics derived
  automatically from language corpora contain human-like biases}.
\newblock \emph{Science}, 356(6334):183--186.

\bibitem[{Cao and Daum{\'e}~III(2020)}]{cao2019toward}
Yang~Trista Cao and Hal Daum{\'e}~III. 2020.
\newblock \href {https://doi.org/10.18653/v1/2020.acl-main.418} {Toward
  gender-inclusive coreference resolution}.
\newblock In \emph{Proceedings of the 58th Annual Meeting of the Association
  for Computational Linguistics}, pages 4568--4595, Online. Association for
  Computational Linguistics.

\bibitem[{Crawford(2017)}]{crawford2017}
Kate Crawford. 2017.
\newblock The trouble with bias.
\newblock keynote talk at Neural Information Processing Systems (NIPS ‘17).

\bibitem[{Escud{\'e}~Font and
  Costa-juss{\`a}(2019)}]{font-costa-jussa-2019-equalizing}
Joel Escud{\'e}~Font and Marta~R. Costa-juss{\`a}. 2019.
\newblock \href {https://doi.org/10.18653/v1/W19-3821} {Equalizing gender bias
  in neural machine translation with word embeddings techniques}.
\newblock In \emph{Proceedings of the First Workshop on Gender Bias in Natural
  Language Processing}, pages 147--154, Florence, Italy. Association for
  Computational Linguistics.

\bibitem[{Hardt et~al.(2016)Hardt, Price, and Srebro}]{hardt2016equality}
Moritz Hardt, Eric Price, and Nathan Srebro. 2016.
\newblock \href {https://dl.acm.org/doi/10.5555/3157382.3157469} {Equality of
  opportunity in supervised learning}.
\newblock In \emph{Proceedings of the 30th International Conference on Neural
  Information Processing Systems}, NIPS'16, page 3323–3331, Red Hook, NY,
  USA. Curran Associates Inc.

\bibitem[{Kocmi et~al.(2020)Kocmi, Limisiewicz, and
  Stanovsky}]{kocmi-etal-2020-gender}
Tom Kocmi, Tomasz Limisiewicz, and Gabriel Stanovsky. 2020.
\newblock \href {https://aclanthology.org/2020.wmt-1.39} {Gender coreference
  and bias evaluation at {WMT} 2020}.
\newblock In \emph{Proceedings of the Fifth Conference on Machine Translation},
  pages 357--364, Online. Association for Computational Linguistics.

\bibitem[{Kumar et~al.(2019)Kumar, Wintner, Smith, and Tsvetkov}]{kumar2019}
Sachin Kumar, Shuly Wintner, Noah~A. Smith, and Yulia Tsvetkov. 2019.
\newblock \href {http://arxiv.org/abs/1909.00453} {Topics to avoid: Demoting
  latent confounds in text classification}.
\newblock \emph{CoRR}, abs/1909.00453.

\bibitem[{Lu et~al.(2018)Lu, Mardziel, Wu, Amancharla, and
  Datta}]{lu2018gender}
Kaiji Lu, Piotr Mardziel, Fangjing Wu, Preetam Amancharla, and Anupam Datta.
  2018.
\newblock \href {https://arxiv.org/pdf/1807.11714.pdf} {Gender bias in neural
  natural language processing}.
\newblock \emph{arXiv:1807.11714}.

\bibitem[{May et~al.(2019)May, Wang, Bordia, Bowman, and
  Rudinger}]{may2019measuring}
Chandler May, Alex Wang, Shikha Bordia, Samuel~R. Bowman, and Rachel Rudinger.
  2019.
\newblock \href {https://doi.org/10.18653/v1/N19-1063} {On measuring social
  biases in sentence encoders}.
\newblock In \emph{Proceedings of the 2019 Conference of the North {A}merican
  Chapter of the Association for Computational Linguistics: Human Language
  Technologies, Volume 1 (Long and Short Papers)}, pages 622--628, Minneapolis,
  Minnesota. Association for Computational Linguistics.

\bibitem[{Menegatti and Rubini(2017)}]{menegatti2017gender}
Michela Menegatti and Monica Rubini. 2017.
\newblock \href {https://doi.org/10.1093/acrefore/9780190228613.013.470}
  {Gender bias and sexism in language}.
\newblock In \emph{Oxford Research Encyclopedia of Communication}.

\bibitem[{Raffel et~al.(2020)Raffel, Shazeer, Roberts, Lee, Narang, Matena,
  Zhou, Li, and Liu}]{raffel2019exploring}
Colin Raffel, Noam Shazeer, Adam Roberts, Katherine Lee, Sharan Narang, Michael
  Matena, Yanqi Zhou, Wei Li, and Peter~J. Liu. 2020.
\newblock \href {http://jmlr.org/papers/v21/20-074.html} {Exploring the limits
  of transfer learning with a unified text-to-text transformer}.
\newblock \emph{Journal of Machine Learning Research}, 21(140):1--67.

\bibitem[{Rudinger et~al.(2018)Rudinger, Naradowsky, Leonard, and
  Van~Durme}]{rudinger2018gender}
Rachel Rudinger, Jason Naradowsky, Brian Leonard, and Benjamin Van~Durme. 2018.
\newblock \href {https://doi.org/10.18653/v1/N18-2002} {Gender bias in
  coreference resolution}.
\newblock In \emph{Proceedings of the 2018 Conference of the North {A}merican
  Chapter of the Association for Computational Linguistics: Human Language
  Technologies, Volume 2 (Short Papers)}, pages 8--14, New Orleans, Louisiana.
  Association for Computational Linguistics.

\bibitem[{Saunders and Byrne(2020)}]{saunders2020reducing}
Danielle Saunders and Bill Byrne. 2020.
\newblock \href {http://arxiv.org/abs/2004.04498} {Reducing gender bias in
  neural machine translation as a domain adaptation problem}.

\bibitem[{Saunders et~al.(2020)Saunders, Sallis, and
  Byrne}]{saunders-etal-2020-neural}
Danielle Saunders, Rosie Sallis, and Bill Byrne. 2020.
\newblock \href {https://aclanthology.org/2020.gebnlp-1.4} {Neural machine
  translation doesn{'}t translate gender coreference right unless you make it}.
\newblock In \emph{Proceedings of the Second Workshop on Gender Bias in Natural
  Language Processing}, pages 35--43, Barcelona, Spain (Online). Association
  for Computational Linguistics.

\bibitem[{Stafanovi{\v{c}}s et~al.(2020)Stafanovi{\v{c}}s, Bergmanis, and
  Pinnis}]{stafanovics-etal-2020-mitigating}
Art{\=u}rs Stafanovi{\v{c}}s, Toms Bergmanis, and M{\=a}rcis Pinnis. 2020.
\newblock \href {https://aclanthology.org/2020.wmt-1.73} {Mitigating gender
  bias in machine translation with target gender annotations}.
\newblock In \emph{Proceedings of the Fifth Conference on Machine Translation},
  pages 629--638, Online. Association for Computational Linguistics.

\bibitem[{Stanovsky et~al.(2019)Stanovsky, Smith, and
  Zettlemoyer}]{stanovsky2019evaluating}
Gabriel Stanovsky, Noah~A. Smith, and Luke Zettlemoyer. 2019.
\newblock \href {https://doi.org/10.18653/v1/P19-1164} {Evaluating gender bias
  in machine translation}.
\newblock In \emph{Proceedings of the 57th Annual Meeting of the Association
  for Computational Linguistics}, pages 1679--1684, Florence, Italy.
  Association for Computational Linguistics.

\bibitem[{Vaccaro et~al.(2019)Vaccaro, Karahalios, Mulligan, Kluttz, and
  Hirsch}]{vaccaro2019contestability}
Kristen Vaccaro, Karrie Karahalios, Deirdre~K. Mulligan, Daniel Kluttz, and Tad
  Hirsch. 2019.
\newblock \href {https://doi.org/10.1145/3311957.3359435} {Contestability in
  algorithmic systems}.
\newblock In \emph{Conference Companion Publication of the 2019 on Computer
  Supported Cooperative Work and Social Computing}, CSCW '19, page 523–527,
  New York, NY, USA. Association for Computing Machinery.

\bibitem[{Vanmassenhove et~al.(2018)Vanmassenhove, Hardmeier, and
  Way}]{vanmassenhove2019getting}
Eva Vanmassenhove, Christian Hardmeier, and Andy Way. 2018.
\newblock \href {https://doi.org/10.18653/v1/D18-1334} {Getting gender right in
  neural machine translation}.
\newblock In \emph{Proceedings of the 2018 Conference on Empirical Methods in
  Natural Language Processing}, pages 3003--3008, Brussels, Belgium.
  Association for Computational Linguistics.

\bibitem[{Xia et~al.(2020)Xia, Field, and Tsvetkov}]{xia2020demoting}
Mengzhou Xia, Anjalie Field, and Yulia Tsvetkov. 2020.
\newblock \href {https://doi.org/10.18653/v1/2020.socialnlp-1.2} {Demoting
  racial bias in hate speech detection}.
\newblock In \emph{Proceedings of the Eighth International Workshop on Natural
  Language Processing for Social Media}, pages 7--14, Online. Association for
  Computational Linguistics.

\bibitem[{Zhang et~al.(2018)Zhang, Lemoine, and Mitchell}]{zhang2018mitigating}
Brian~Hu Zhang, Blake Lemoine, and Margaret Mitchell. 2018.
\newblock \href {https://doi.org/10.1145/3278721.3278779} {Mitigating unwanted
  biases with adversarial learning}.
\newblock In \emph{Proceedings of the 2018 AAAI/ACM Conference on AI, Ethics,
  and Society}, AIES '18, page 335–340, New York, NY, USA. Association for
  Computing Machinery.

\bibitem[{Zhao et~al.(2017)Zhao, Wang, Yatskar, Ordonez, and
  Chang}]{zhao2017men}
Jieyu Zhao, Tianlu Wang, Mark Yatskar, Vicente Ordonez, and Kai-Wei Chang.
  2017.
\newblock \href {https://doi.org/10.18653/v1/D17-1323} {Men also like shopping:
  Reducing gender bias amplification using corpus-level constraints}.
\newblock In \emph{Proceedings of the 2017 Conference on Empirical Methods in
  Natural Language Processing}, pages 2979--2989, Copenhagen, Denmark.
  Association for Computational Linguistics.

\bibitem[{Zmigrod et~al.(2019)Zmigrod, Mielke, Wallach, and
  Cotterell}]{zmigrod-etal-2019-counterfactual}
Ran Zmigrod, Sabrina~J. Mielke, Hanna Wallach, and Ryan Cotterell. 2019.
\newblock \href {https://doi.org/10.18653/v1/P19-1161} {Counterfactual data
  augmentation for mitigating gender stereotypes in languages with rich
  morphology}.
\newblock In \emph{Proceedings of the 57th Annual Meeting of the Association
  for Computational Linguistics}, pages 1651--1661, Florence, Italy.
  Association for Computational Linguistics.

\end{thebibliography}
\bibliographystyle{acl_natbib}

\appendix

\section{Ethics and Data Statement}
\label{app:data}
The WMT-2014 training data used for the study comes from was chosen because this dataset, as used in the 2014 ACL Workshop in Statistical Machine Translation, was originally used to train T5 for machine translation; therefore, training on this data could be reasonably expected to provide similar results as those on the original model. The data used here comes from parallel English-French and English-German texts. The English data consists primarily of General American English and British English; other dialects of English, such as African-American English, are underrepresented in this corpus. Similarly, the French and German data consists primarily of varieties of these languages as spoken in Europe. The training set therefore cannot not provide a balanced representation of the various dialects and speaker demographics of these languages.

The WinoMT dataset used for evaluation (see Section \ref{section:results}) was chosen because it is the predominant benchmark for evaluating gender bias in machine translation. It consists of English-language templates in General American English that were then translated into other languages. Thus, an important avenue for further research is to examine whether bias mitigation techniques like this one are effective on more translation pairs and language varieties.

Finally, we acknowledge that our approach assumes a binary notion of gender and does not account for other gender identities; we recommend that future work explore avenues for gender-inclusive translation as well.

\section{Implementation Details}
\label{app:implementation}
 The number of training epochs was manually fine-tuned from 1 to 5 on the development set; all models presented here were trained for 2 epochs, except for the EN-FR gender direction model (1 epoch) and EN-DE baseline (3 epochs). A learning rate of 0.001 was used for both the translation model and the adversary and the Adam optimizer was used for all models. We used T5-base (220 million parameters); training time for each model varied from 24 to 72 hours on one NVIDIA Quadro RTX 8000 GPU. 
 
 The development set BLEU scores corresponding to the final models used here were:
 
 \begin{table}[h]
\resizebox{0.4\textwidth}{!}{
\begin{tabular}{Sl Sl Sl}
\hline
    & & \textbf{BLEU} \\
    \hline
\multirow{3}{*}{EN-FR} &\textbf{Baseline} & 30.6  \\
&\textbf{Gender direction method} & 29.3 \\
&\textbf{Pronoun usage method}    & 33.8 \\
\hline
\multirow{3}{*}{EN-DE} &\textbf{Baseline} & 28.4 \\
&\textbf{Gender direction method} & 31.3 \\
&\textbf{Pronoun usage method}    & 30.3 \\
\hline
\end{tabular}}. 
\label{table:devset}
\end{table}



\end{document}